\documentclass[conference]{IEEEtran}
\IEEEoverridecommandlockouts
\usepackage[bottom]{footmisc}
\usepackage{amsmath,amssymb,amsfonts}
\usepackage{algorithmic}
\usepackage[tablename=Table]{caption}
\usepackage{rotating}
\usepackage{svg}
\usepackage{dirtytalk}
\usepackage[colorinlistoftodos,prependcaption,textsize=normal]{todonotes}
\usepackage[capitalise]{cleveref}
\usepackage{enumitem}
\usepackage{array}
\usepackage{arydshln}
\usepackage{lipsum}
\usepackage{float}
\usepackage[figurename=Figure]{caption}
\usepackage{graphicx}
\usepackage[backend=bibtex,style=ieee,natbib=true]{biblatex}
\usepackage{subfig}
\usepackage{mwe}
\usepackage{graphics}

\usepackage[autostyle=true]{csquotes} 
\usepackage{tabularx,ragged2e,booktabs,caption}
\usepackage{textcomp}
\renewcommand\IEEEkeywordsname{Keywords}
\makeatletter
\renewcommand{\fnum@figure}{Figure. \thefigure}
\makeatother
\begin{document}
\bstctlcite{IEEEexample:BSTcontrol}
\title{Surrogate Assisted Methods for the Parameterisation of Agent-Based Models}


\newcommand\copyrighttext{%
  \footnotesize \textcopyright 2020 IEEE. Personal use of this material is permitted.  Permission from IEEE must be obtained for all other uses, in any current or future media, including reprinting/republishing this material for advertising or promotional
  purposes, creating new collective works, for resale or redistribution to servers or lists, or reuse of any copyrighted component of this work in other works.}
\newcommand\arxivcopyrightnotice{%
\begin{tikzpicture}[remember picture,overlay]
\node[anchor=south,yshift=10pt] at (current page.south) {\fbox{\parbox{\dimexpr\textwidth-\fboxsep-\fboxrule\relax}{\copyrighttext}}};
\end{tikzpicture}%
}
\def\BibTeX{{\rm B\kern-.05em{\sc i\kern-.025em b}\kern-.08em
    T\kern-.1667em\lower.7ex\hbox{E}\kern-.125emX}}
\author{\IEEEauthorblockN{Rylan Perumal}
\IEEEauthorblockA{\textit{School of Computer Science and Applied Mathematics} \\
\textit{University of the Witwatersrand}, South Africa \\
rylan.perumal@wits.ac.za}
\and
\IEEEauthorblockN{Terence L van Zyl}
\IEEEauthorblockA{\textit{Institute for Intelligent Systems} \\
\textit{University of Johannesburg}, South Africa \\
tvanzyl@uj.ac.za}
}
\newcommand{\code}{\texttt}
\newcolumntype{P}[1]{>{\centering\arraybackslash}m{#1}}
\maketitle
\arxivcopyrightnotice
\begin{abstract}
Parameter calibration is a major challenge in agent-based modelling and simulation (ABMS). As the complexity of agent-based models (ABMs) increase, the number of parameters required to be calibrated grows. This leads to the ABMS equivalent of the \say{curse of dimensionality}. We propose an ABMS framework which facilitates the effective integration of different sampling methods and surrogate models (SMs) in order to evaluate how these strategies affect parameter calibration and exploration. We show that surrogate assisted methods perform better than the standard sampling methods. In addition, we show that the XGBoost and Decision Tree SMs are most optimal overall with regards to our analysis. 
\end{abstract}

\begin{IEEEkeywords}
Agent-based modelling and simulation, surrogate models, infectious disease epidemiology, machine learning
\end{IEEEkeywords}
\section{INTRODUCTION}
Agent-based models (ABMs) offer the possibility to model many complex real-world scenarios~\cite{abm_use_case}. These scenarios range from modelling the spread of an epidemic within a population to modelling trends based on agents' behaviour in the stock market. With the increase in model complexity, the number of parameters that need to be calibrated, to allow the model to match real-world data, grows~\cite{dengue}. As a result, searching for meaningful parameter combinations can become computationally prohibitive. Machine learning (ML) models, namely surrogate models (SMs), are capable of effectively searching the parameter space of ABMs. Previously, SMs have been used to classify whether a candidate parameter vector is a good parameter calibration to the ABM to match the real-world data \cite{surrogate_og}.

\begin{figure}[htb!]
    \hspace*{1.5cm}\includegraphics[scale=0.29]{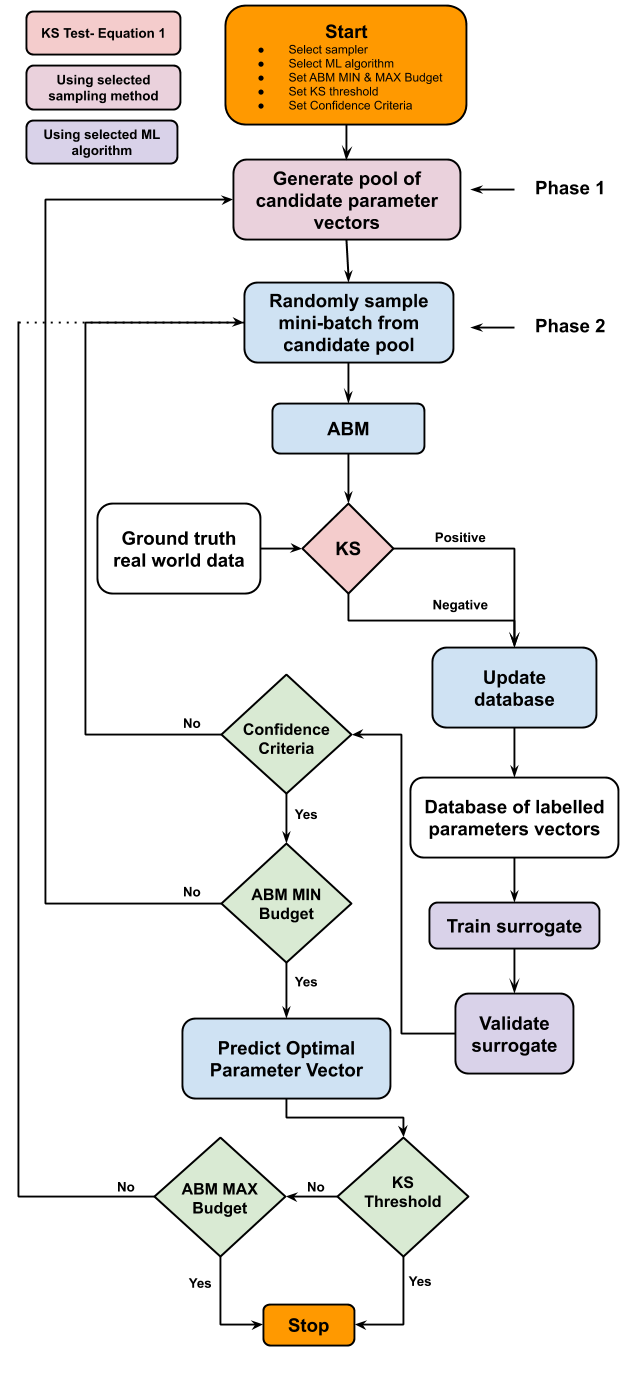}
    \caption{Agent-based modelling and simulation (ABMS) framework}
    \label{fig:ABMS_framework}
\end{figure}

In this paper we present the results obtained using our implementation of the agent-based modelling and simulation (ABMS) framework shown in Figure~\ref{fig:ABMS_framework} to facilitate the parameterisation challenges of ABMs. In addition, we compare surrogate assisted sampling methods using different SMs. Our results obtained show:
\begin{itemize}
    \item XGBoost and DT SMs perform the best at assisting the parameterisation of ABMs.
    \item The surrogate assisted method \textit{XGBoost Random} is is able to get within 98.5\% of the optimal distribution with the lowest number of mini-batch evaluations.
    \item Overall we show that surrogate assisted methods are more likely to estimate the most optimal parameter vector which generates a synthetic data distribution that matches the real data distribution.
    \item We note the difficulty of calibrating ABMs when one considers that we are trying to estimate only seven of the possible nine parameters using synthetic data and raise the need for further investigation for using real-world data.
\end{itemize}

\section{BACKGROUND} \label{background}
\subsection{Agent-Based Modelling and Simulation in Epidemiology}
Agent-based modelling and simulation (ABMS) is an effective technique that is a natural fit for modelling infectious diseases. Agent-based models (ABMs) are capable of modelling interactions between individuals and their environment. Further, they are capable of capturing unexpected emergent patterns and trends during an epidemic, which are the result of combined individual agent behaviours. Each agent within an ABM is provided with different agent characteristics. The agents act autonomously, governed by the interaction of their set of pre-defined rules and distinctive characteristics. As a result, ABMs require sufficient complexity to model almost any complex real-world scenario \cite{abm_use_case}. In addition, ABMS can be used as a substitute for a real-world epidemiological study since it is often not feasible or even possible to run a real-world experiment \citep{abm_methods_techniques, taxonomy, public_health_abm}. 

While there are significant benefits to using ABMS for infectious disease epidemiology, there are equally limitations. ABMs ordinarily require long run times due to the increased computational complexity resulting from agent interactions incorporated into the ABM. Additionally, model validation and parameterisation present major challenges in ABMS, specifically in reflecting real-world data. Of these two challenges, we are particularly interested in the parameterisation of ABMs. Difficulty with finding correct parameterisations of ABMs lead to extensive calibration efforts resulting in increased model development time. As more complexity is added to the model, the parameter space expands, leading to the ABMS equivalent of the \say{curse of dimensionality} problem. The result is impractical memory and computational costs when searching for meaningful parameter combinations \citep{surrogate_og, bayesian_abm, dengue, influenza_review, surrogate_improve, taxonomy}.

\subsection{Surrogate Models in Agent-Based Modelling and Simulation}
One approach to overcoming the computational limitations of ABMS is the use of surrogate models (SMs). SMs are machine learning (ML) models that act as functional approximations to ABMs. In addition, SMs provide a computationally tractable solution to addressing the issues of parameter sensitivity analysis, robust analysis and empirical validation in ABMS. These properties make SMs appealing for approximating significantly complex ABMs that are computationally expensive to validate and calibrate \cite{van_deHoog_surrogate, surrogate_improve, surrogate_og}.

Previously, Gaussian process regression, also known as the Kriging method, has been used as a surrogate modelling approach to facilitate parameter space exploration and sensitivity analysis challenges in ABMS. However, Kriging's performance is dependent on the model's ability to estimate the true spatial continuity of the data~\citep{surrogate_og, surrogate_improve, bargigli_kriging, dosi_kriging}.

\citet{surrogate_og} presented an alternate approach, overcoming some of the limitations of the Kriging method. An iterative algorithm is proposed for training a SM to effectively approximate the ABM. The novel approach is realised by combining ML and intelligent iterative sampling. It is demonstrated that a model's parameter space can be effectively searched utilising fewer computational resources adopting their approach. In their work the XGBoost non-parametric ML algorithm is used, where the SM is built in a stage-wise fashion, allowing optimisation of an arbitrary differentiable loss function. This method was applied to the Asset Pricing Model by \citet{asset_pricing} and the Island Growth model by \citet{island_growth}. The results obtained show that the SM is an accurate functional approximation of the ABM. Further, the SM radically reduced the computation time for large-scale parameter space calibration and exploration. \citet{surrogate_improve} improve on the work of \citet{surrogate_og} using the CatBoost ML algorithm. They show the surrogate is able to approximate the ABM and as a result further reduce parameter calibration and exploration time.

\subsection{Quasi-Random Sobol Sampling}
\citet{sobol} proposed the Quasi-Random Sobol sampling strategy, used by \citet{surrogate_og} and \citet{surrogate_improve}, which belongs to the class of Quasi-Monte Carlo methods. In particular, this approach is ideal when sampling from distributions with an unknown topology such as the parameter space of ABMs for infectious diseases. This approach is able to guarantee uniformity of a distribution by filling up spaces with random points.

\section{METHODOLOGY FOR AGENT-BASED MODELLING AND SIMULATION} \label{methodology}

\subsection{Agent-Based Modelling and Simulation Framework}

Figure \ref{fig:ABMS_framework} is a representation of our implemented ABMS framework inspired by the algorithm of \citet{surrogate_og}. Our framework starts by selecting the initial configuration. This includes the sampling method, the ML algorithm, the values for the \textit{ABM MIN Budget} and the \textit{ABM MAX Budget} limits, a threshold value for the KS test statistic and the confidence criteria. We generate a pool of candidate parameter vectors utilising the sampling method. Subsequently, a mini-batch of these vectors is randomly sampled, where each candidate vector is used as input for the ABM. The ABM generates synthetic epidemic data based on the input parameter vector. We compare the similarity between the distribution of the synthetic epidemic data to the real-world data using Equation \ref{equation1} and the corresponding parameter vector is labelled accordingly. Once we have labelled all of the parameter vectors from the mini-batch, we subsequently include these in a ground-truth database. The ground-truth database contains all the parameter vectors we have evaluated and labelled. We employ the ground-truth database to construct an SM using ML techniques and then evaluate if the SM meets the confidence criteria. We either check if we have reached the \textit{ABM MIN Budget} limit or go to \textit{Phase 2}. After examining the \textit{ABM MIN Budget} limit, we either resume at \textit{Phase 1} or predict the optimal parameter vector from our database of labelled vectors. We then verify to see if the predicted parameter vector's KS test statistic value is $\leq$ the set \textit{KS Threshold}. Depending on the outcome of the check, we either evaluate the \textit{ABM MAX Budget} limit and go to \textit{Phase 2} or stop.

\subsection{Agent-based model}
The agent-based model (ABM) used in this framework is a pre-existing model which has been implemented by the \textit{Julia} library, \textit{Agents.jl}\footnote{https://juliadynamics.github.io/Agents.jl/stable/models/}. The ABM used is a continuous space virus spread model, where the disease transmission dynamics follows the basic Susceptible-Infected-Recovered (SIR) framework, proposed by \citet{kermack1927_sir}, simulated for $2000$ time-steps. The SIR framework models the ratio of susceptible, infected and recovered individuals within a population. Table \ref{table1} contains the true input parameter values of the ABM that we have considered for parameterisation. Parameters $1, 2, 3, 6$ and $7$ are sampled between the range $(0, 1)$  and parameters $4$ and $5$ are sampled between the range $(0, 41)$ days. We have chosen the number $41$, which equates to $1000$ time steps (half of the total) in our ABM, as our upper bound to allow the sampling methods to generate candidate parameter vectors where the \textit{Infection Period} and \textit{Detection Time} are greater than true value.

\begin{table}[b]
\begin{minipage}{.5\textwidth}
\centering
\caption{Table showing the true values of the considered ABM's parameters.}
\begin{tabular}{|l|c|}
\hline
\multicolumn{1}{|c|}{\textbf{Parameters}} & \multicolumn{1}{|c|}{\textbf{True Values}} \\ \hline
1. Transmission Probability ($\beta$)                                      & 0.639                                                             \\ \hline
2.  Reinfection Probability                            & 0.129                                                             \\ \hline
3. Death Probability                                                     & 0.44                                                              \\ \hline
4. Infection Period                                             & 30 days                                                             \\ \hline
5. Detection Time                                                          & 14 days                                                            \\ \hline
6. Speed                                             & 0.002                                                             \\ \hline
7. Interaction Radius                                                         & 0.012                                                           \\ \hline
\end{tabular}
\label{table1}
\end{minipage}
\end{table}

\subsection{Kolmogorov-Smirnov Test}
The two-sample Kolmogorov-Smirnov (KS) test is used to compare the distributions of the simulated data to the real data as follows:
\begin{equation}
    D_{R, S} = \sup_{x} \large| F_{R}(x) - F_{S}(x) \large|, \label{equation1}
\end{equation}
where $x$ represents the feature we are measuring (number of infected individuals) and $F_{R}$ and $F_{S}$ are the distribution functions of the real and simulated data respectively \cite{surrogate_og}. The KS test requires the cumulative distributions of the samples being compared to be calculated. The $99\%$ critical value is used to reject the null hypothesis (two distributions are the same). The corresponding parameter vector is labelled as negative if the two distributions are not the same and positive if the two distributions are similar as seen in Figure~\ref{fig:ABMS_framework}. In addition, the data distributions that we are comparing are time series. The empirical KS test is formulated to assess the distance between two independent and identically distributed samples. To use the KS test on our time-series data, we convert the time-series to a cumulative distribution function by first performing a cumulative sum along the time dimension. We then scale the cumulative sum so as to arrive at a cumulative distribution which maintains the integrity of the time-series. This makes our problem scale-invariant to the \textit{population size} simulated by the ABM, which allows us to measure the similarity of the exact epidemic trends between the two time-series distributions.
\subsection{Sampling Methods}
\subsubsection{Random Sampler}
This sampling technique is used as our baseline. We generate $N$ random candidate parameter vectors. The length of all the candidate parameter vectors generated are dependent on the number of parameters we are parameterising.

\subsubsection{Surrogate Assisted Random Sampler}
To improve the accuracy and efficiency of the random sampler, we propose the surrogate assisted random sampler. Once we have a confident SM, we then re-generate a new set of random candidate parameter vectors. The new set is passed as input to the SM which classifies the candidates as positive or negative parameter calibrations. The parameter pool is then re-initialised using an $\epsilon$-greedy algorithm. We select positively predicted candidates $90\%$ of the time and selecting negatively predicted candidates $10\%$ of the time. This encourages exploration of the parameter space. This will drive our sampling method to mostly generate candidate parameters vectors which it classifies as positive calibrations.

\subsubsection{Quasi-Random Sobol Sampler}
We have used the \textit{Sobol.jl}\footnote{https://github.com/stevengj/Sobol.jl} package from the \textit{Julia} library which is based on the works of \citet{sobol_original} and \citet{sobol_remark}. This sampling method generates low-discrepancy-sequences of points that are equally distributed around an $N$-dimensional hyper-cube. 
 
\subsubsection{Surrogate Assisted Quasi-Random Sobol Sampler}
In addition, we propose the surrogate assisted quasi-random sobol sampler. This sampling method functions in the same way as the surrogate assisted random sampler. The only difference between the two methods, other than the SM used, is that one generates pseudo-random samples and the other quasi-random sobol samples.

\subsection{Surrogate Models}
We evaluated the following machine learning algorithms for learning surrogate models:
\begin{itemize}
    \item \textbf{eXtreme Gradient Boosting (XGBoost):} A decision tree based machine learning algorithm, which is based off the framework by \citet{friedman}, that is scalable and efficient in its implementation \cite{xgboost}.
    \item \textbf{Decision Tree (DT):} A classification algorithm which works off a tree-like structure, where the nodes represent a feature/attribute and the branches represent the decision rule which leads to the outcome of that decision \cite{dt}. 
    \item \textbf{Support Vector Machine (SVM):} A classification algorithm which finds a hyperplane in an $N$-dimensional space in order to differentiate between different classes of data points \cite{svm}.
\end{itemize}
The ground-truth database of labelled parameter vectors is used at each iteration to train a SM. We split the database into a training and validation set using an $80/20$ split. The machine learning algorithm is trained using the training set and the \textit{F1 Score} is evaluated on the validation set. The confidence of the SM, measured by accuracy on the validation set, should increase as additional examples are stored in the ground-truth database.

\subsection{Sanity Check}
To ensure that our implemented ABMS framework in Figure \ref{fig:ABMS_framework} is reliable, we need to conduct a sanity check. Given a set of known optimal parameters, $\theta^{*}$, we use this as input for the constructed ABM. The ABM will then generate a synthetic dataset based on the value of $\theta^{*}$ as output. We subsequently use the synthetic dataset as the ground truth real-world data in Figure \ref{fig:ABMS_framework}. Thereafter we run through the ABMS framework and observe if we are able to approximate $\theta^{*}$.

\subsection{Experiment Setup}
\subsubsection{Initial ABMS Configurations}
The following initial configurations are set as: \textit{ABM MIN Budget} $= 500$, \textit{ABM MAX Budget} $= 2500$, \textit{Batch Size} $= 50$ and \textit{KS Threshold} $= 0.005$ ($\approx 99.5\%$ similar to the true distribution). Lastly, the confidence criteria of the SM is set so that we have at least evaluated a proportional number of candidate parameter vectors, dependent on the \textit{Batch Size} and the number of parameters we are parameterising, at a validation \textit{F1 Score} $\geq 0.90$. We conducted a total of $56$ experiments, averaging each experiment $10$ times, where for each average the true parameters were varied. We compared all of the sampling methods implemented, parametersing parameters $1, \dots, 7$ as seen in Table \ref{table1}.


\subsubsection{Hardware Specifications}
The machine used to run our experiments consisted of an Intel Xeon CPU E5-2683 v4 @ 2.10GHz processor with 64 CPUs and 256GB of RAM using the Ubuntu 18.04.4 LTS operating system.
\section{RESULTS}
\begin{figure}[b]
    \centering
    {
    \includegraphics[scale=0.30]{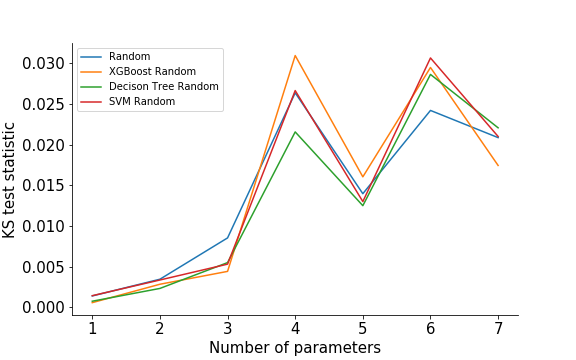}
    }\\{
    \includegraphics[scale=0.30]{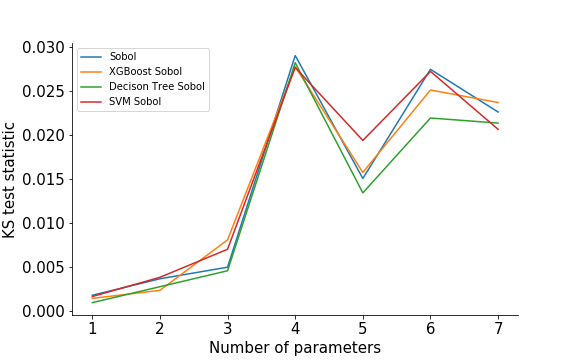}
    }
    \caption{Comparison between the KS test statistic values of the different sampling methods implemented and the number of parameters being estimated.}
    \label{fig2}
\end{figure}

We evaluate the performance of the various surrogate models, random sampler and quasi-random sobol sampler in the context of the following metrics:
\begin{itemize}
    \item \textbf{L$_2$ Norm:} The euclidean distance between the true input parameter vectors and the optimal predicted parameter vectors.
    \item \textbf{KS Test Statistic:} The maximum distance between two empirical time-series cumulative distributions functions. The similarity of the time-series distributions increases as this value tends to $0$.
    \item \textbf{Mini-batch Evaluations to Success (BMS):} Success is defined as quality criterion that needs to be achieved i.e. a solution within 99\% or 95\% of the known optimal parameter values. BMS is the number of mini-batches the framework required to reach success, i.e. the BMS at a success of 99\% would be the number of mini-batches used to get to within a 99\% of the optimal value.
\end{itemize}
Table \ref{table2} contains standardised the L$_{2}$ Norm and KS test statistic values for each of the methods that we have implemented, calibrating parameters $1, \dots, 7$. Figure \ref{fig2} shows the results of KS test statistic values seen in Table \ref{table2}. The results we have obtained clearly illustrate that trying to parameterise more than $3$ parameters results in a gradual decrease in performance as measured by the KS test statistic. 
\\
The top of Figure \ref{fig2} shows that \textit{Decision Tree Random} consistently outperforms \textit{Random} for parameters $1$, $2$, $3$, $4$, $5$ and $7$. \textit{XGBoost Random} outperforms \textit{Random} for parameters $1$, $2$, $3$ and $7$. The bottom of Figure \ref{fig2} shows for parameterising parameters $1, \dots, 7$, that \textit{Decision Tree Sobol} outperforms \textit{Sobol} for all parameters. The results from Table \ref{table2} and Figure \ref{fig2} clearly illustrate that XGBoost and DT surrogates are able to get the lowest KS test statistic values and also the lowest L$_2$ norm values overall.
\begin{table*}[t!]
\caption{Standardised $\text{L}_2$ Norm and KS test statistic values for the optimal predicted parameter vectors using each of the sampling methods and surrogate models implemented.}
\label{table2}
 \begin{center}
 \resizebox{2\columnwidth}{!}{
\begin{tabular}{|c|c|c|c|c|c|c|c||c|c|c|c|c|c|c|}
\hline
\textbf{}               & \multicolumn{7}{c||}{\textbf{Standardised $\text{L}_2$ Norm}}               & \multicolumn{7}{c|}{\textbf{KS Test Statistic}}                \\ \hline \hline
\textbf{Parameters}     & \textbf{1} & \textbf{2} & \textbf{3} & \textbf{4} & \textbf{5} & \textbf{6} & \textbf{7}& \textbf{1} & \textbf{2} & \textbf{3} & \textbf{4} & \textbf{5} & \textbf{6} & \textbf{7}  \\ \hline

\textbf{Random}          &  0.009417 &  0.092984 &  0.243392 &  0.900970 &  \textbf{0.738482} &  1.019350 &  1.618652 &
0.001403 &  0.003458 &  0.008536 &  0.026368 &  0.013968 &  0.024202 &  0.020861   \\ \hline

\textbf{XGBoost Random} &  \textbf{0.008793} &  0.079441 &  0.203277 &  0.719190 &  0.971193 &  1.092223 &  1.654796 &
\textbf{0.000579} &  0.002841 &  \textbf{0.004421} &  0.030946 &  0.016041 &  0.029479 &  \textbf{0.017441}   \\ \hline

\textbf{DT Random}     &  0.016872 &  \textbf{0.076023} &  0.244910 & \textbf{ 0.630211} &  0.934826 &  1.622811 &  1.460234 
&  0.000755 &  \textbf{0.002326} &  0.005501 & \textbf{0.021563} &  \textbf{0.012496} &  0.028628 &  0.022066 \\ \hline

\textbf{SVM Random}      &  0.021645 &  0.141421 &  0.259465 &  0.808132 &  9.889844 &  1.139878 &  1.325837 
&  0.001435 &  0.003363 &  0.005310 &  0.026643 &  0.013002 &  0.030652 &  0.021000  \\ \hline

\textbf{Sobol}          &  0.020650 &  0.154425 &  0.229348 &  0.636288 &  1.080810 &  1.073848 &  \textbf{1.087858}  
&  0.001793 &  0.003663 &  0.004975 &  0.028976 &  0.015037 &  0.027414 &  0.022571   \\ \hline

\textbf{XGBoost Sobol}  &  0.016922 &  0.083839 &  0.219330 &  0.552427 &  0.877853 &  \textbf{1.005900} &  1.692152
&  0.001437 &  0.002350 &  0.008076 &  0.027716 &  0.015726 &  0.025069 &  0.023649   \\ \hline

\textbf{DT Sobol}       &  0.014734 &  0.120076 &  \textbf{0.145425} &  0.689795 &  0.748668 &  1.182660 &  1.397635
&  0.000956 &  0.002763 &  0.004571 &  0.028165 &  0.013404 &  \textbf{0.021895} &  0.021317   \\ \hline

\textbf{SVM Sobol}      &  0.013012 &  0.133107 &  0.267395 &  0.695961 &  0.975068 &  1.495472 &  1.118029 
&  0.001650 &  0.003829 &  0.007011 &  0.027626 &  0.019353 &  0.027181 &  0.020611    \\ \hline

\end{tabular}}
\end{center}
\end{table*}

Table \ref{table3} contains the averaged minimum number of mini-batch evaluations it took to reach the optimal predicted parameter vector. The table also details the number of mini-batch evaluations to succeed at different percentage intervals for parameterising $7$ of the ABM's parameters.  \textit{XGBoost Sobol} is able to reach success at 97\% and 97.5\% in only $14.3$ and $20$ mini-batch evaluations on average respectively. \textit{XGBoost Random} and \textit{SVM Sobol} are able to reach success at 98\% and 98.5\% respectively in the least amount of mini-batch evaluations on average. As we scale the problem size and the complexity of the ABM, we may find that doing so many mini-batch evaluations is computationally infeasible to reach that level of success.

\begin{table}[htb!]
\caption{Number of mini-batch evaluations on average to success at (97\%, 97.5\%, 98\% and 98.5\%) for estimating $7$ ABM parameters.}
\label{table3}
\resizebox{\columnwidth}{!}{
\begin{tabular}{|l|c|c|c|c|}
\hline
                          & \multicolumn{4}{c|}{\textbf{Mini-batch Evaluations}}                                                                 \\ \hline\hline
\textbf{Sampling Methods} & \textbf{97\%} & \textbf{97.5\%} & \textbf{98\%} & \textbf{98.5\%}  \\ \hline
\textbf{Random}          &  15.6 & 25.4 &  34.4 &  43.0  \\ \hline
\textbf{XGBoost Random} &  17.0 & 25.6 &  32.9 &  \textbf{39.1} \\ \hline
\textbf{DT Random}        &  23.0 & 26.6 &  30.2 &  40.9 \\ \hline
\textbf{SVM Random}       &  22.4 & 27.3 &  31.6 &  40.1  \\ \hline
\textbf{Sobol}            &  20.9 & 26.8 &  35.7 &  40.5    \\ \hline
\textbf{XGBoost Sobol}   & \textbf{14.3} & \textbf{20.0} &  37.2 &  41.1  \\ \hline
\textbf{DT Sobol} &  24.5 & 25.2 &  29.2 &  40.5   \\ \hline
\textbf{SVM Sobol}  &  20.5 & 27.7 &  \textbf{27.7} &  43.1  \\ \hline
\end{tabular}}
\end{table}

\section{CONCLUSION} \label{conclusion}
We have implemented an ABMS framework which allowed us to effectively swap out and replace alternative sampling methods and surrogate models. In addition, the framework allows us to evaluate the performance of parameter calibration and exploration challenges in ABMS. To be specific, we are interested in utilising our approach for infectious disease epidemiology. Our results demonstrate to us that the surrogate assisted methods perform much better than the \textit{Random} and \textit{Quasi-Random Sobol} sampling methods. In addition, employing an XGBoost and DT surrogate is most optimal at assisting the sampling methods with regards to approximating the real data distribution. Our future work aims to improve on our ABMS framework but include different aspects of optimisation and to evaluate more intelligent sampling methods and ML algorithms to learn SMs.

\printbibliography[heading=bibintoc]

@article{dengue,
  title={Why should we apply ABM for decision analysis for infectious diseases?—An example for dengue interventions},
  author={Miksch, Florian and Jahn, Beate and Espinosa, Kurt Junshean and Chhatwal, Jagpreet and Siebert, Uwe and Popper, Nikolas},
  journal={PloS one},
  volume={14},
  number={8},
  year={2019},
  publisher={Public Library of Science}
}

@article{taxonomy,
  title={A taxonomy for agent-based models in human infectious disease epidemiology},
  author={Hunter, Elizabeth and Mac Namee, Brian and Kelleher, John D},
  journal={Journal of Artificial Societies and Social Simulation},
  volume={20},
  number={3},
  year={2017},
  publisher={JASSS}
}

@article{influenza_review,
  title={Modelling the global spread of diseases: A review of current practice and capability},
  author={Walters, Caroline E and Mesl{\'e}, Margaux MI and Hall, Ian M},
  journal={Epidemics},
  volume={25},
  pages={1--8},
  year={2018},
  publisher={Elsevier}
}

@article{kermack1927_sir,
  title={A contribution to the mathematical theory of epidemics},
  author={Kermack, William Ogilvy and McKendrick, Anderson G},
  journal={Proceedings of the royal society of london. Series A, Containing papers of a mathematical and physical character},
  volume={115},
  number={772},
  pages={700--721},
  year={1927},
  publisher={The Royal Society London}
}

@article {abm_methods_techniques,
	author = {Bonabeau, Eric},
	title = {Agent-based modeling: Methods and techniques for simulating human systems},
	volume = {99},
	number = {suppl 3},
	pages = {7280--7287},
	year = {2002},
	doi = {10.1073/pnas.082080899},
	publisher = {National Academy of Sciences},
	journal = {Proceedings of the National Academy of Sciences}
}

@article{public_health_abm,
  title={Agent-based modeling in public health: current applications and future directions},
  author={Tracy, Melissa and Cerd{\'a}, Magdalena and Keyes, Katherine M},
  journal={Annual review of public health},
  volume={39},
  pages={77--94},
  year={2018},
  publisher={Annual Reviews}
}

@INPROCEEDINGS{abm_use_case,  author={C. M. {Macal} and M. J. {North}},  booktitle={Proceedings of the 2009 Winter Simulation Conference (WSC)},   title={Agent-based modeling and simulation},   year={2009},  volume={},  number={},  pages={86-98}, doi={10.1109/WSC.2009.5429318.}}

@article{surrogate_improve,
  title={Validation and Calibration of an Agent-Based Model: A Surrogate Approach},
  author={Zhang, Yi and Li, Zhe and Zhang, Yongchao},
  journal={Discrete Dynamics in Nature and Society},
  volume={2020},
  year={2020},
  publisher={Hindawi}
}

@article{surrogate_og,
  title={Agent-based model calibration using machine learning surrogates},
  author={Lamperti, Francesco and Roventini, Andrea and Sani, Amir},
  journal={Journal of Economic Dynamics and Control},
  volume={90},
  pages={366--389},
  year={2018},
  publisher={Elsevier}
}

@article{van_deHoog_surrogate,
  title={Surrogate modelling in (and of) agent-based models: A prospectus},
  author={van der Hoog, Sander},
  journal={Computational Economics},
  volume={53},
  number={3},
  pages={1245--1263},
  year={2019},
  publisher={Springer}
}

@article{bayesian_abm,
  title={Bayesian estimation of agent-based models},
  author={Grazzini, Jakob and Richiardi, Matteo G and Tsionas, Mike},
  journal={Journal of Economic Dynamics and Control},
  volume={77},
  pages={26--47},
  year={2017},
  publisher={Elsevier}
}

@article{bargigli_kriging,
  title={Network calibration and metamodeling of a financial accelerator agent based model},
  author={Bargigli, Leonardo and Riccetti, Luca and Russo, Alberto and Gallegati, Mauro},
  journal={Journal of Economic Interaction and Coordination},
  pages={1--28},
  year={2018},
  publisher={Springer}
}

@article{dosi_kriging,
  title={The effects of labour market reforms upon unemployment and income inequalities: an agent-based model},
  author={Dosi, Giovanni and Pereira, Marcelo C and Roventini, Andrea and Virgillito, Maria Enrica},
  journal={Socio-Economic Review},
  volume={16},
  number={4},
  pages={687--720},
  year={2018},
  publisher={Oxford University Press}
}

@article{asset_pricing,
  title={Heterogeneous beliefs and routes to chaos in a simple asset pricing model},
  author={Brock, William A and Hommes, Cars H},
  journal={Journal of Economic dynamics and Control},
  volume={22},
  number={8-9},
  pages={1235--1274},
  year={1998},
  publisher={Elsevier}
}

@article{island_growth,
  title={Exploitation, exploration and innovation in a model of endogenous growth with locally interacting agents},
  author={Fagiolo, Giorgio and Dosi, Giovanni},
  journal={Structural Change and Economic Dynamics},
  volume={14},
  number={3},
  pages={237--273},
  year={2003},
  publisher={Elsevier}
}

@article{sobol_original,
author = {Bratley, Paul and Fox, Bennett L.},
title = {Algorithm 659: Implementing Sobol’s Quasirandom Sequence Generator},
year = {1988},
issue_date = {March 1988},
publisher = {Association for Computing Machinery},
address = {New York, NY, USA},
volume = {14},
number = {1},
doi = {10.1145/42288.214372},
journal = {ACM Trans. Math. Softw.},
month = mar,
pages = {88–100},
numpages = {13}
}

@article{sobol_remark,
author = {Joe, Stephen and Kuo, Frances Y.},
title = {Remark on Algorithm 659: Implementing Sobol’s Quasirandom Sequence Generator},
year = {2003},
issue_date = {March 2003},
publisher = {Association for Computing Machinery},
address = {New York, NY, USA},
volume = {29},
number = {1},
doi = {10.1145/641876.641879},
journal = {ACM Trans. Math. Softw.},
month = mar,
pages = {49–57},
numpages = {9}
}

@article{sobol,
  title={Quasi-random sequences and their discrepancies},
  author={Morokoff, William J and Caflisch, Russel E},
  journal={SIAM Journal on Scientific Computing},
  volume={15},
  number={6},
  pages={1251--1279},
  year={1994},
  publisher={SIAM}
}

@book{svm,
  title={The nature of statistical learning theory},
  author={Vapnik, Vladimir},
  year={2013},
  publisher={Springer science \& business media}
}

@article{dt,
  title={The decision tree classifier: Design and potential},
  author={Swain, Philip H and Hauska, Hans},
  journal={IEEE Transactions on Geoscience Electronics},
  volume={15},
  number={3},
  pages={142--147},
  year={1977},
  publisher={IEEE}
}

@article{xgboost,
  title={Xgboost: extreme gradient boosting},
  author={Chen, Tianqi and He, Tong and Benesty, Michael and Khotilovich, Vadim and Tang, Yuan},
  journal={R package version 0.4-2},
  pages={1--4},
  year={2015}
}

@article{friedman,
  title={Greedy function approximation: a gradient boosting machine},
  author={Friedman, Jerome H},
  journal={Annals of statistics},
  pages={1189--1232},
  year={2001},
  publisher={JSTOR}
}

\end{document}